\documentclass[10pt, a4paper]{article}
\usepackage{lrec}
\pdfoutput=1
\usepackage{graphicx}
\usepackage{tabularx}
\usepackage{multirow}
\usepackage{makecell}
\usepackage{booktabs}
\usepackage{floatrow}
\usepackage{subfigure}
\usepackage{array}
\usepackage{soul}

\usepackage[outdir=./]{epstopdf}
\usepackage[utf8]{inputenc}
\usepackage[encapsulated]{CJK}

\usepackage{hyperref}
\usepackage{xstring}

\usepackage{amsmath,amssymb}
\usepackage{enumerate}

\usepackage{verbatim}
\usepackage{tikz}

\title{\textbf{Shallow Discourse Annotation for Chinese TED Talks}}

\name{Wanqiu Long\textsuperscript{1,†}, Xinyi Cai\textsuperscript{1,†}, James E. M. Reid\textsuperscript{2}, Bonnie Webber\textsuperscript{2}, Deyi Xiong\textsuperscript{1,*}}

\address{\textsuperscript{1}College of Intelligence and Computing, Tianjin University, Tianjin \\ \textsuperscript{2}School of Informatics, University of Edinburgh, Edinburgh \\
         longwanqiu01@163.com, \{xinyicai,dyxiong\}tju.edu.cn\\
         jamesemreid@icould.com,bonnie@inf.ed.ac.uk\\}

\abstract{
 Text corpora annotated with language-related properties are an important resource for the development of Language Technology. The current work contributes a new resource for Chinese Language Technology and for Chinese-English translation, in the form of a set of TED talks (some originally given in English, some in Chinese) that have been annotated with discourse relations in the style of the Penn Discourse TreeBank, adapted to properties of Chinese text that are not present in English. The resource is currently unique in annotating discourse-level properties of planned spoken monologues rather than of written text. An inter-annotator agreement study demonstrates that the annotation scheme is able to achieve highly reliable results. \\ \newline  }

\begin{document}

\maketitleabstract

\section{Introduction}

\renewcommand{\thefootnote}{}
\footnote{Wanqiu Long and Xinyi Cai have contributed equally to this
work. Corresponding author: Deyi Xiong, dyxiong@tju.edu.cn}

Researchers have recognized that performance improvements in natural language processing (NLP) tasks such as summarization \cite{jones2007automatic}, question answering
\cite{verberne2007evaluating}, and machine translation \cite{guzman2014using} can come from recognizing discourse-level properties of text. These include properties such as the how new entities are introduced into the text, how entities are subsequently referenced (e.g., coreference chains), and how clauses and sentences relate to one another. Corpora in which such properties have been manually annotated by experts can be used as training data for such tasks, or seed data for creating additional ``silver annotated'' data. Penn Discourse Treebank (PDTB), a lexically grounded method for annotation, is a shallow approach to discourse structure which can be adapted to different genres. Annotating discourse relations both within and across sentences, it aims to have wide application in the field of natural language processing. PDTB can effectively help extract discourse semantic features, thus serving as a useful substrate for the development and evaluation of neural models in many downstream NLP applications.\\

Few Chinese corpora are both annotated for discourse properties and publicly available. The available annotated texts are primarily newspaper articles. The work described here annotates another type of text – the planned monologues found in TED talks, following the annotation style used in the Penn Discourse TreeBank, but adapted to take account of properties of Chinese described in Section 3.\\

TED talks (TED is short for technology, entertainment, design), as examples of planned monologues delivered to a live audience \cite{greenbaum1996comparing}, are scrupulously translated to various languages. Although TED talks have been annotated for discourse relations in several languages \cite{zeyrek2018multilingual}, this is the first attempt to annotate TED talks in Chinese (either translated into Chinese, or presented in Chinese), providing data on features of Chinese spoken discourse. Our annotation by and large follows the annotation scheme in the PDTB-3, adapted to features of Chinese spoken discourse described below. \\

The rest of the paper is organized as follows: in Section 2, we review the related existing discourse annotation work. In Section 3, we briefly introduce PDTB-3 \cite{webber2019penn} and our adapted annotation scheme by examples. In Section 4, we elaborate our annotation process and the results of our inteannotator-agreement study. Finally, in Section 5, we display the results of our annotation and preliminarily analyze corpus statistics, which we compare to the relation distribution of the CUHK Discourse TreeBank for Chinese. (CUHK-DTBC)\cite{zhou2014cuhk}.

\section{Related work}

Following the release of the Penn Discourse Treebank (PDTB-2) in 2008 \cite{prasad2008penn}, several remarkable Chinese discourse corpora have since adapted the PDTB framework \cite{prasad2014reflections}, including the Chinese Discourse Treebank \cite{zhou2012pdtb}, HIT Chinese Discourse Treebank (HIT-CDTB) \newcite{zhou2014cuhk}, and the Discourse Treebank for Chinese (DTBC) \cite{zhou2014cuhk}. Specifically, Xue proposed the Chinese Discourse Treebank (CDTB) Project \cite{xue2005penn}. From their annotation work, they discussed the matters such as features of Chinese discourse connectives, definition and scope of arguments, and senses disambiguation, and they argued that determining the argument scope is the most challenging part of the annotation. To further promote their research, \newcite{zhou2012pdtb} presented a PDTB-style discourse corpus for Chinese. They also discussed the key characteristics of Chinese text which differs from English, e.g., the parallel connectives, comma-delimited intra-sentential implicit relations etc. Their data set contains 98 documents from the Chinese Treebank \cite{xue2005penn}. In 2015, Zhou and Xue expanded their corpus to 164 documents, with more than 5000 relations being annotated. \newcite{huang-chen-2011-chinese} constructed a Chinese discourse corpus with 81 articles. They adopted the top-level senses from PDTB sense hierarchy and focused on the annotation of inter-sentential discourse relations. \newcite{zhang2014chinese} analyzed the differences between Chinese and English, and then presented a new Chinese discourse relation hierarchy based on the PDTB system, in which the discourse relations are divided into 6 types: temporal, causal, condition, comparison, expansion and conjunction. And they constructed a Chinese Discourse Relation corpus called HIT-CDTB based on this hierarchy. Then, \newcite{zhou2014cuhk} presented the first open discourse treebank for Chinese, the CUHK Discourse Treebank for Chinese. They adapted the annotation scheme of Penn Discourse Treebank 2 (PDTB-2) to Chinese language and made adjustments to 3 aspects according to the previous study of Chinese linguistics. However, they just reannotated the documents of the Chinese Treebank and did not annotate inter-sentence level discourse relations. \\
\setcounter{footnote}{0}
\renewcommand{\thefootnote}{\arabic{footnote}}

It is worth noting that, all these corpora display a similar unbalanced distribution that is likely to be associated with them all being limited to text from the same NEWS genre. In particular, these two senses (Expansion and Conjunction) represent \ 80 \% of the relations annotated in the CDTB\footnote{CDTB uses a flat set of senses in which Conjunction and Expansion are distinct.
}. \\

In addition, although annotating spoken TED talks has been done on other several languages before \cite{zeyrek2018multilingual}, to our knowledge, there is no recent annotation work for Chinese spoken discourses, or particularly for Chinese Ted talks. However, there is some evidence that noticeable differences in the use of discourse connectives and discourse relations can be found between written and spoken discourses \cite{rehbein-etal-2016-annotating}. Here, by using the new PDTB-3 sense hierarchy and annotator, which has not been used for Chinese annotation before, we annotated Chinese Ted talks to help others be aware of the differences between the Chinese discourse structure of written and spoken texts and will make our corpus publicly available to benefit the discourse-level NLP researches for spoken discourses.

\section{PDTB and our Annotation Scheme}

The annotation scheme we adopted in this work is based on the framework of PDTB, incorporating the most recent PDTB (PDTB-3) relational taxonomy and sense hierarchy \cite{webber2019penn}, shown in Table 1. PDTB follows a lexically grounded approach to the representation of discourse relations \cite{miltsakaki2004penn}. Discourse relations are taken to hold between two abstract object arguments, named Arg1 and Arg2 using syntactic conventions, and are triggered either by explicit connectives or, otherwise, by adjacency between clauses and sentences. As we can see from Table 1, the PDTB-3 sense hierarchy has 4 top-level senses (Expansion, Temporal, Contingency, Contrast) and second- and third-level senses for some cases. With obvious differences ranging from the conventions used in annotation, to differences in senses hierarchy, PDTB-3 gives rigorous attention to achieving as much consistency as possible while annotating discourse relations.\\




\begin{table*}[ht!]
\centering
\subfigure{
\begin{tabular}[t]{|l|l|l|}
\hline
\multirow{3}{*}{Temporal} & Synchronous                   & --         \\ \cline{2-3} 
                          & \multirow{2}{*}{Asynchronous} & Precedence \\ \cline{3-3} 
                          &                               & Succession \\ \hline
\end{tabular}
}

\quad
\subfigure{
\begin{minipage}[b]{0.475\linewidth}
\centering

\begin{tabular}{|l|l|l|}
\hline
\multirow{10}{*}{Contingency} & \multirow{3}{*}{Cause} & Reason \\ \cline{3-3}
                              &        				   & Result \\ \cline{3-3}
                              &						   & Negative-result \\ \cline{2-3}
                              & \multirow{2}{*}{Condition} & Arg1-as-cond \\ \cline{3-3}
                              & 					   & Arg2-as-cond \\ \cline{2-3}
                              & \multirow{2}{*}{Negative condition}     & Arg1-as-negcond \\ \cline{3-3}
                              &						   & Arg2-as-negcond \\ \cline{2-3}
                              & \multirow{3}{*}{Purpose} & Arg1-as-goal \\ \cline{3-3}
                              &							 & Arg2-as-goal \\ \cline{3-3}
                              & 						 & Arg2-as-negGoal \\ \hline \hline
\multirow{4}{*}{Comparison} & Contrast 				& --				\\ \cline{2-3}
							& Similarity			& --				\\ \cline{2-3}
							& \multirow{2}{*}{Concession} & Arg1-as-denier	\\ \cline{3-3}
							&							  & Arg2-as-denier	\\ \hline
\end{tabular}
\end{minipage}
}%
\subfigure{
\begin{minipage}[b]{0.475\linewidth}
\centering
\begin{tabular}{|l|l|l|}
\hline
\multirow{13}{*}{Expansion} & Conjunction & --  		\\ \cline{2-3}
							& Disjunction & --			\\ \cline{2-3}
							& Equivalence & --			\\ \cline{2-3}
							& \multirow{2}{*}{Instantiation}	& Arg1-as-instance	\\ \cline{3-3}
							&									& Arg2-as-instance 	\\ \cline{2-3}
							& \multirow{2}{*}{Level-of-detail}	& Arg1-as-detail	\\ \cline{3-3}
							& 									& Arg2-as-detail 	\\ \cline{2-3}
							& \multirow{2}{*}{Substitution}		& Arg1-as-subst		\\ \cline{3-3}
							&									& Arg2-as-subst		\\ \cline{2-3}
							& \multirow{2}{*}{Execption}		& Arg1-as-excpt		\\ \cline{3-3}
							&									& Arg2-as-excpt		\\ \cline{2-3}
							& \multirow{2}{*}{Manner}			& Arg1-as-manner	\\ \cline{3-3}
							& 									& Arg2-as-manner	\\ \hline

\end{tabular}
\end{minipage}
}
\caption{PDTB-3 Sense Hierarchy \protect\cite{webber2019penn}}
\end{table*}

Previously, all Chinese annotation work using PDTB style followed the settings of PDTB-2. Some researchers tried to adapt it in lines of the Chinese characteristics. For example, \newcite{zhou2012pdtb} annotated the parallel connectives continuously rather than discontinuously due to the greater use of parallel connectives in Chinese and a reduced use of explicit connectives in general. \newcite{zhou2014cuhk} added some additional senses into the hierarchy. However, PDTB-3, as a new and enriched version, not only has paid greater attention to intra-sentential senses, but also has incorporated some of those additional senses. Therefore, we just made several modifications including removing, adding, or disambiguating for the practical use of PDTB-3 into our Chinese annotation.\\

In practice, using the PDTB annotator tool, we annotated an explicit connective, identified its two arguments in which the connective occurs, and then labeled the sense. For implicit relations, when we inferred the type of relation between two arguments, we tried to insert a connective for this relation, and also the inserted connective is not so strictly restricted, extending to expressions that can convey the sense of the arguments. If a connective conveys more than one sense or more than one relation can be inferred, multiple senses would be assigned to the token. Our adaptations towards PDTB-3 will be introduced from the perspectives of arguments, relations and senses as follows.

\subsection{Arguments}

The argument-labelling conventions used in the PDTB-2 had to be modified to deal with the wider variety of discourse relations that needed to be annotated consistently within sentences in the PDTB-3. In particular, in labelling intra-sentential discourse relations, a distinction was made between relations whose arguments were in coordinating syntactic structures and ones whose arguments were in subordinating syntactic structures. For coordinating structures, arguments were labelled by position (Arg1 first, then Arg2), while for subordinating structures, the argument in subordinate position was labelled Arg2, and the other, Arg1, independent of position.\\

For discourse in Chinese, this can introduce an unwanted ambiguity. Example 1 is a typical example for illustrate this phenomenon. In the examples throughout the paper, explicit connectives are underlined, while implicit Discourse Connectives and the lexicalizing expression for Alternative Lexicalizations are shown in parentheses and square brackets respectively. The position of the arguments is indicated by the attached composite labels to the right square brackets, and the relation lables and sense lables can be seen in the parentheses at the end of arguments. When the arguments, relations or senses are ambiguous, there may be no corresponding labels shown in the examples.

\begin{CJK}{UTF8}{gbsn}
\begin{enumerate}[(1)]
\item  \underline{因为} \@ \@ \@ \  你  \@ \@  让 \;\;\;\, 我 \;生气，   \underline{所以}，我 \,要让 \\
\@\@ Because  you  make  me  angry, so \;\;\;\;\;\; I \;\; want   \\
你\qquad 更难过。(Explicit, Cause.Result) \\
you to be sadder.\\
``You made me angry, so I return it double back.''

\end{enumerate}

While ``because'' and ``so'' are rarely found together as connectives in a sentence in English, it is not uncommon to find them used concurrently as a paired connective in Chinese. Therefore, due to this difference, the annotators tend to have no idea about which clause is subordinate. Therefore, if we regard the first clause as subordinating structure and “因 为”(because)as connective, then the sense would be Contingency.Cause.Reason. By contrast, the sense would be Contingency.Cause.Result, when the second clause is regarded as Arg2. To get rid of this kind of ambiguity, we just take the first as Arg1 and the second Arg2 regardless of the fact that the parallel connectives are surbodinating or coordinating.

\subsection{Relations}

There are two new types of relation in PDTB-3: AltlexC and Hypophora. Hypophora is an explicitly marked question-response pairs, first used in annotating the TED- MDB \cite{zeyrek2018multilingual}. In Hypophora relations, Arg1 expresses a question and Arg2 offers an answer, with no explicit or implicit connective being annotated (Example 2). Because of the nature of TED talks, many relations in both the TED-MDB and in our Chinese TED talks are examples of ``Hypophora''. However, not all discourse relations whose first argument is a question are Hypophora. Example 3, instead of seeking information and giving answer, is just a rhetorical question expressing negation by imposing a dramatic effect.\\

\begin{enumerate}
\item[(2)] [我到底\qquad 要\qquad \quad 讲\qquad \quad \; 什么Arg1]?\\
\@\@ \@\@ I \; on earth am going to talk about what ?\\
 \@\@ [最后\quad 我\;决定\qquad 要\; 讲\qquad 教育Arg2]。(Hypophora) \\
\@\@ \qquad Finally,  I \;decided to  talk about education .\\
``What am I gonna say? Finally, I decided to talk about education.''
\end{enumerate}

\begin{enumerate}
\item[(3)]他\;说 ： “ 我\; 是\quad 三 天\quad \; 一 小 哭 、 \qquad 五 天\\
\@\@ He said, ``  I \; am  three days a little cry, five days\\
 一 大 哭 。 ” 这样\quad \quad\; 你 有\quad 比较\; 健康 吗 ？\\ 
 a\; lot cry.'' In this way, you are more\; healthier?\\
 都 是\quad 悲伤 ，\quad 并 不 是 每 一 个 人 ，\;每 一 次\\
 All\; are \; sadness, not\qquad everyone,\quad every time \\
  感受 到 悲伤 的 时候 ，都 一定 会 流泪 、\quad 甚至 大哭 。\\
  feel\quad sad 's time, \quad\; would\; shed tears \;even\quad cry.\\
``He said, ``Three times I cry a little, and five times I cry a lot.''
Is that healthier?
Everyone gets sad, but that's not to say that whenever someone feels sad, they necessarily will cry.''

\end{enumerate}

In addition, we found a new issue when identifying Hypophora, which is shown in Example 4. In this example, we have a series of questions, rather than a series of assertions or a question-response pair. We attempted to capture the rhetorical links by taking advantage of our current inventory of discourse relations. Here, two implicit relations were annotated for this example, and the senses are Arg2-as-detail and Result+SpeechAct respectively. Therefore, when there are subsequent texts related to a question or a sequence of questions, we would not just annotated them as Hypophora but had to do such analysis as what we did for the examples shown.\\

\begin{enumerate}
\item[(4)][情绪 ，\; 它 到底\quad 是 什么Rel1-Arg1 ]? (\underline{具体来说})  \\
\@\@ Emotion, it on earth is what?\qquad \qquad \quad (Specially) \\
\@\@[它 是 好还是 不 好Rel1-Arg2,Rel2-Arg1]？(Implicit，Arg2-as-detail)\\
\@\@ It is good or bad? \\
 (\underline{所以})[你 会 想要 拥有 它 吗Rel2-Arg2]？ (Implicit，Result+SpeechAct)\\
 (So)\quad You want to have it ?\\
``What is it exactly?
Is it good or bad?
Do you want to have them?''
\end{enumerate}

Besides, it is widely accepted that the ellipsis of subject or object are frequently seen in Chinese. Then for EntRel, if facing this situation where one of the entities in Arg1 or Arg2 is omitted, we still need to annotate this as EntRel (Example 5). In this following example, we can see in Arg2, the pronoun which means ``that'' is omitted, but in fact which refers to the phenomenon mentioned in Arg1, so here there is still an EntRel relation between this pair of arguments.

\begin{enumerate}
\item[(5)][我们会以讽刺的口吻来谈论，\; \quad 并且会 \\
\@\@ We in ironic terms talk about, \quad \quad \quad
and \\
\@\@加上引号 : \;“进步”Arg1 ]  \\
\@\@add quotes: ``Progress''.\\
\@\@ [我想是有原因的，\; \quad 我们也知道\quad 是\quad 什么\quad \\
\@\@I think there are reasons, \;we also know are what \\
\@\@ 原因Arg2]。(EntRel)\\
\@\@ reasons.\\
``We talk about it in ironic terms with little quotes
around it:``Progress''.Okay, there are reasons for that, and I think we know what those reasons are.''
\end{enumerate}

\subsection{Senses}

The great improvement in the sense hierarchy in PDTB-3 enables us to capture more senses with additional types and assign the senses more clearly. For example, the senses under the category of Expansion such as level of detail, manner, disjunction and similarity are indispensable to our annotation. Therefore, we nearly adopted the sense hierarchy in PDTB-3, just with few adaptations. On the one hand, we removed the third level sense ``Negative condition+SpeechAct'', since it was not used to label anything in the corpus. On the other hand, we added the Level-2 sense ``Expansion.Progression''. This type of sense applies when Arg1 and Arg2 are coordinating structure with different emphasis. The first argument is annotated as Arg1 and the second as Arg2. This sense is usually conveyed by such typical connectives as “不 但 (not only)... 而 且 (but also)...”, “甚 至 (even)... 何 况 (let alone)...”,“... 更 (even more)...”(Example 6).\\

\begin{enumerate}
  \item[(6)][我 去 了\; 聋人 俱乐部 ，观看 了\; 聋人 的 \\
  \@\@ I went to   deaf  clubs,  \quad\, saw  the deaf person's\\
  \@\@表演 Arg1]。[我\underline{甚至}\,去 了\,田纳西州 纳什维尔的 \\
  performances. I \quad even went to the Nashville's \\
  “ 美国 聋人 小姐 ” \hspace{3em}\, 选秀赛Arg2]。(Explicit, Progression.Arg2-as-progr)\\
   ``the Miss Deaf'' America contest.\\
   ``I went to deaf clubs. I saw performances of deaf theater and of deaf poetry. I even went to the Miss Deaf America contest in Nashville.''
\end{enumerate}

Another issue about sense is the inconsistency when we annotated the implicit relations. \newcite{zhou2012pdtb} did not insert connective for implicit relations, but we did this for further researches with regard to implicit relations. However, we found that in some cases where different connectives can be inserted into the same arguments to express the same relation, the annotators found themselves in a dilemma. We can see that Example 7 and Example 8 respectively insert ``so'' and ``because'' into the arguments between which there is a causal relation, but the senses in these two examples would be Cause.Result and Cause.Reason. The scheme we adopted for this is that we only take the connectives that we would insert into account, and the position and sense relations of arguments would depend on the inserted connectives.\\

\begin{enumerate}
\item[(7)][“克服\hspace{3em}\; 逆境 ”\quad  这一说法\; 对\;我\\
\@\@“Overcome the adversity”  this  phrase for me\\
\@\@根本 \qquad \,不\quad 成立 Arg1]，（所以）[别人  让\quad\; 我  \\
completely not justified,\quad (so)\quad  others asked me\\
\@\@ 就 这一话题\, 说\quad 几 句    的时候， 我很不自在Arg2]。(Implicit, Cause.Result) \\
to this topic talk about some,\qquad I felt uneasy.\\
\item[(8)]（因为）[“克服\hspace{3em}\quad 逆境 ”\quad\; 这一说法\\
(Because) “overcome the adversity”  this  phrase\\
\@\@对\,我来说根本\qquad\; 不\quad 成立Arg2],\quad [ 别人\\
for me \quad completely not justified, \qquad (so)  others\\
\@\@让\quad\, 我就\quad 这一话题\; 说几句的时候，\;我很\\
 asked me to this topic,  talk about some, \quad\, I felt\\
 \@\@不自在Arg1]。(Implicit, Cause.Reason) \\
   uneasy.\\
  ````overcome the adversity'' this phrase never sat right with me, and I always felt uneasy trying to answer people's questions about it.''
\end{enumerate}

\section{Annotation Procedure}

In this section, we describe our annotation process in creating the Chinese TED discourse treebank. To ensure annotation quality, the whole annotation process has three stages: annotators training, annotation, post-annotation. The training process intends to improve the annotators’ annotation ability, while after the formal annotation, the annotated work was carefully checked by the supervisor, and the possible errors and inconsistencies were dealt with through discussions and further study. 

\subsection{Annotator training}

The annotator team consists of a professor as the supervisor, an experienced annotator and a researcher of  PDTB as counselors, two master degree candidates as annotators. Both of the annotators have a certain theoretical foundation of linguistics. To guarantee annotation quality, the annotators were trained through the following steps: firstly, the annotators read the PDTB-3 annotation manual, the PDTB-2 annotation manual and also other related papers carefully; next, the annotators tried to independently annotate same texts, finding out their own uncertainties or problems respectively and discussing these issues together; then, the annotators were asked to create sample annotations on TED talks transcripts for each sense from the top level to the third. They discussed the annotations with the researchers of the team and tried to settle disputes. When sample annotations are created, this part of process is completed; based on the manuals, previous annotation work and also the annotators’ own pre-annotation work, they made a Chinese tutorial on PDTB guidelines, in which major difficulties and perplexities, such as the position and the span of the arguments, the insert of connectives, and the distinction of different categories of relations and senses, are explained clearly in detail by typical samples. This Chinese tutorial is beneficial for those who want to carry out similar Chinese annotation, so we made this useful tutorial available to those who want to carry out similar annotation\footnote{available at:\href{https://github.com/tjunlp-lab/Shallow-Discourse-Annotation-for-Chinese-TED-Talks}{https://github.com/tjunlp-lab/Shallow-Discourse-Annotation-for-Chinese-TED-Talks} \label{web}}; finally, to guarantee annotation consistency, the annotators were required to repeat their annotation-discussion process until their annotation results show the Kappa value \textgreater 0.8 for each of the indicators for agreement.
\begin{table}[]
	\centering
	
	\begin{tabular}{lccc}
\hline
&\textbf{TED talks (No.)}&\textbf{words}&\textbf{Relations}\\
		\hline
\multirow{8}*{Original}&1&3082&195\\   

~&3&4079&250\\

~&5&5145&326\\

~&6&4993&343\\
~&7&4282&284\\
~&9&2220&132\\
~&10&2549&168\\
~&12&2614&172\\
\hline
\multirow{8}*{Translated}&769&5617&243\\   

~&824&5710&335\\

~&837&1961&99\\

~&1756&6004&340\\
~&1971&1083&50\\
~&1978&3025&158\\
~&2009&1307&50\\
~&2150&1636&67\\	
\hline
	\end{tabular}
	\caption{The length and the number of relations of each text }
\end{table}

\subsection{Corpus building}

At present, our corpus has been released publicly\textsuperscript{\ref{web}}. Our corpus consists of two parts with equal number of texts: (1) 8 English TED talks translated into Chinese, just like the talks in the TED-MDB, all of which were originally presented in English and translated into other languages (including German, Lithuanian, Portuguese,Polish, Russian and Turkish) \cite{zeyrek2018multilingual}. (2) 8 Chinese TED talks originally presented in Taipei and translated into English. We got the texts by means of extracting Chinese and English subtitles from TED talks videos \footnote{available at:\href{https://www.youtube.com/results?search\_query=TED+taipei}{https://www.youtube.com/results?search\_query=\\TED+taipei}}. Firstly, we just annotated the talks given in English and translated in Chinese. But after considering the possible divergencies between translated texts and the original texts, we did our annotation for the Taipei TED talks, which were delivered in Chinese. The parallel English texts are also being annotated for discourse relations, but they are not ready for carrying out a systematic comparison between them. At the current stage, we annotated 3212 relations for the TED talks transcripts with 55307 words，and the length of each talk (in words) and the number of annotated 
\begin{table}[!th]
\begin{center}
\begin{tabular}{l  l}

      \hline
      \textbf{Relation type} & \\
      \hline
      Agreement & 0.95\\
      Kappa & 0.93\\
      \hline
      \textbf{Senses ( Top level )} & \\
      \hline
      Agreement & 0.94\\
      Kappa & 0.92\\
      \hline
      \textbf{Senses ( Second level )} & \\
      \hline
      Agreement & 0.85\\
      Kappa & 0.83\\
      \hline
      \textbf{Senses ( Third level )} & \\
      \hline
      Agreement & 0.85\\
      Kappa & 0.83\\
      \hline
      \textbf{Argument order} & \\
      \hline
      Agreement & 0.99\\
      Kappa & 0.98\\
      \hline
      \textbf{Argument scope} & \\
      \hline
      Agreement & 0.88\\
      Kappa & 0.86\\
      \hline
\end{tabular}
\caption{Agreement study}
\end{center}
\end{table}
relations in each talks can be found from Table 2. These TED talks we annotated were prudently selected from dozens of candidate texts. The quality of texts which is principally embodied in content, logic, punctuation and the translation are the major concerns for us. Moreover, when selecting the texts from the Taipei talks, we ruled out those texts which are heavy in dialogues. Some speakers try to interact with the audience, asking the questions, and then commenting on how they have replied. However, what we were annotating was not dialogues. In spite of critically picking over the texts, we still spent considerable time on dealing with them before annotation such as inserting punctuation and correcting the translation. Moreover, before annotation, we did word segmentation by using Stanford Segmenter and corrected improper segmentation. \\

While annotating, we assigned the vast majority of the relations a single sense and a small proportion of relations multiple senses. Unlike previous similar Chinese corpora which primarily or just annotated the relations between sentences, we annotated not only discourse relations between sentences but intra-sentential discourse relations as well. To ensure building a high-quality corpus, the annotators regularly discussed their difficulties and confusions with the researcher and the experienced annotator in the whole process of annotation. After discussion, the annotators reached agreement or retained the differences for few ambiguities.

\subsection{Agreement study}

We measured intra-annotator agreement between two annotators in three aspects: relations, senses, arguments. To be specific, the annotators’ consistency in annotating the type of a specific relation or sense and the position and scope of arguments are measured. To assess the consistency of annotations and also eliminate coincidental annotations, we used agreement rates, which is calculated by dividing the number of senses under each category where the annotators annotate consistently by the total number of each kind of sense. And considering the potential impact of unbalanced distribution of senses, we also used the Kappa value. And the final agreement study was carried out for the first 300 relations in our corpus. We obtained high agreement results and Kappa value for the discourse relation type and top-level senses ($\geq{0.9} $ ). However, what we did was more than this, and we also achieved great results on the second-level and third-level senses for the sake of our self-demand for high-quality, finally achieving agreement of 0.85 and Kappa value of 0.83 for these two deeper levels of senses.\\[-5.5mm]\\

Table 3 also shows that agreement on argument order is almost 1.0 (kappa = 0.99). This means that the guidelines were sufficiently clear that the annotators rarely had difficulty in deciding the location of Arg1 and Arg2 when the senses are determined. Concerning the scope of arguments, which is seen as the most challenging part in the annotation work \cite{xue2005penn}, our agreement and Kappa value on this are 0.88 and 0.86 respectively, while the agreement of the scope of arguments depends on whether the scopes of two arguments the anotators annotated are completely the same. Under such strict requirement, our consistency in this respect is still significantly higher than that of other annotation work done before, for we strictly obeyed the rules of “minimality principle” mentioned in the PDTB-3 annotation manual and got a clearer perspective of supplementary information. Therefore, the annotators are better at excluding the information that do not fall within the scope of the discourse relation.\\[-5.5mm]\\

It is useful to determine where the annotators disagreed most with each other. The three senses where most disagreement occurred are shown in Table 4. The disagreements were primarily in labelling implicit relations. The highest level of disagreement occurred with Expansion.Conjunction and Expansion.Detail, accounting for 12.5 \% among all the inconsistent senses. It is because, more often than not, the annotators failed to judge whether the two arguments make the same contribution with respect to that situation or both arguments describing the same has different level of details. The second highest level of disagreement is reflected in Conjunction and Asynchronous, accounting for 9.3 \%. Besides, Contrast and Concession are two similar senses, which are usually signaled by the same connectives like “但是”, “而”, “不过”, and all these words can be translated into“but”in English. Hence, the annotators sometimes tend to be inconsistent when distinguishing them.\\

\begin{table*}[!th]
\begin{center}
\begin{tabular}{l  c}

      \hline
      \textbf{Cases} & \textbf{Proportion}\\
      \hline
      Expansion.conjunction and Expansion.detail & 12.5\%\\
      Expansion.conjunction and Temporal.asynchronous & 9.3\%\\
      Comparison.Contrast and Comparison.concession & 6.3\%\\
      \hline

\end{tabular}

\caption{ Disagreements between annotators: Percentage of cases}
 \end{center}
\end{table*}

\section{Results}

In regard to discourse relations, there are 3212 relations, of which 1237 are explicit relations (39\%) and 1174 are implicit relation (37\%) (Figure 1). The remaining 801 relations include Hypophora, AltLex, EntRel, and NoRel. Among these 4 kinds of relations, what is worth mentioning is AltLex(Alternative Lexicalizations ),which only constitutes 3\% but is of tremendous significance, for we are able to discover inter- or intra-sentential relations when there is no explicit expressions but AltLex expressions conveying the relations.  but AltLex expressions(eg, 这导致了(this cause), 一个例子是(one example is... ), 原 因 是 (the reason is), etc.). Originally in English, AltLex is supposed to contain both an anaphoric or deictic reference to an actual argument and an indication of the type of sense \cite{DBLP:conf/coling/PrasadJW10}. While for Chinese, the instances of Altlex do not differ significantly from those annotated in English. To prove this, two examples are given as below (Example 9 and Example 10). From our annotation, we realized that Altlex deserves more attention, for which can effectively help to recogonize types of discourse relations automatically.\\
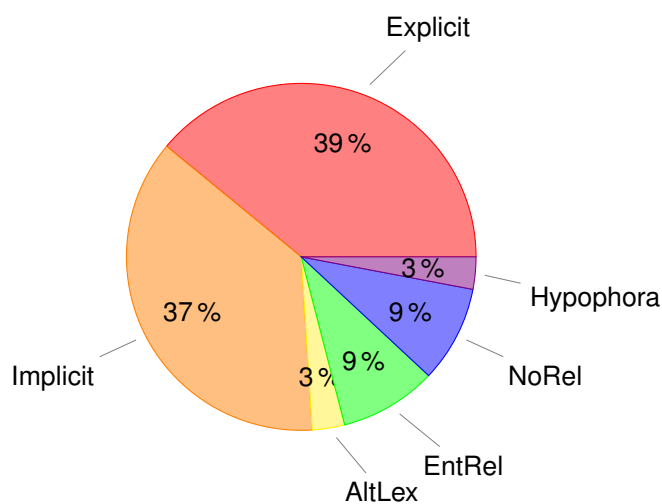
\begin{figure}[!th]
\begin{center}
\def\angle{0}
\def\radius{2.3}
\def\cyclelist{{"red","orange","yellow","green","blue","violet"}}
\newcount\cyclecount \cyclecount=-1
\newcount\ind \ind=-1
\begin{tikzpicture}[nodes = {font=\sffamily}]
  \foreach \percent/\name in {
      39/Explicit,
      37/Implicit,
      3/AltLex,
      9/EntRel,
      9/NoRel,
      3/Hypophora
    } {
      \ifx\percent\empty\else               
        \global\advance\cyclecount by 1     
        \global\advance\ind by 1            
        \ifnum5<\cyclecount                 
          \global\cyclecount=0              
          \global\ind=0                     
        \fi
        \pgfmathparse{\cyclelist[\the\ind]} 
        \edef\color{\pgfmathresult}         
        \draw[fill={\color!50},draw={\color}] (0,0) -- (\angle:\radius)
          arc (\angle:\angle+\percent*3.6:\radius) -- cycle;
        \node at (\angle+0.5*\percent*3.6:0.7*\radius) {\percent\,\%};
        \node[pin=\angle+0.5*\percent*3.6:\name]
          at (\angle+0.5*\percent*3.6:\radius) {};
        \pgfmathparse{\angle+\percent*3.6}  
        \xdef\angle{\pgfmathresult}         
      \fi
    };
\end{tikzpicture}
\caption{Relation distribution}
\label{figure 1}
\end{center}
\end{figure}

\begin{enumerate}
\item[(9)][“国内\quad\quad 许多被截肢者,\;\quad \quad \quad 无法使用  \\
\@\@ in this country, many of the amputees, cannot use\\
\@\@ 他们的假肢Arg1],[这其中的原因是] [他们由于 \\
\@\@ their prostheses,\quad \quad the reason
was \quad their \\ 
\@\@ 假肢接受腔\quad \quad 无法\quad 与残肢\quad 适配 \quad 而\\
\@\@ prosthetic sockets cannot their leg fit well so that \\
\@\@ 感到疼痛Arg2].(AltLex, Cause.Reason) \\
\@\@ felt painful.] \\
“Many of the amputees in the country would not use their prostheses. The reason, I would come to find out, was that their prosthetic sockets were painful because they did not fit well.”
\end{enumerate}

\begin{enumerate}
 \item[(10)]三年级的时候考进秀朗小学的游泳班,\\
in third grade, got in the swimming class at Xiu Lang\\
\@\@[ \qquad  \qquad \qquad \qquad \quad \quad 这个班每天的 \qquad 游泳 \\
elementary school, this class everyday's swimming\\
 训练量高达 3000 米 Arg1], 我发现 [这样的训练量\\
 volumm reach 3000 meters, I realized the training load\\
 使][我 无法同时兼顾两种乐器 Arg2]。(AltLex,\\
 Cause.Result)\\
   make me cannot learn the two instruments at the same time\\
  ``I got in the swimming class at Xiu Lang elementary
    school when I was in third grade. We had to swim up
    to 3000 meters every day. I realized the training load
    was too much for me to learn the two instruments at
    the same time.''
\end{enumerate}

\begin{table}[!th]
\begin{center}
\begin{tabular}{l| c c}
      \hline
      &\textbf{Our courpus} & \textbf{CUHK-DTBC}\\
      \hline
      Comparison & 15.7\% & 11\%\\
      Contingency & 27.6\% & 23\%\\
      Expansion & 37.5\% & 52\%\\
      Temporal & 19.2\% & 14\%\\
      \hline
\end{tabular}
\caption{Distribution of class level senses in our corpus and 400 documents of CUHK-DTBC}
 \end{center}
\end{table}

\begin{table}[!th]
\begin{center}
\begin{tabular}{l c}
      \hline
      Cause & 20\%\\
      \hline
      Conjunction & 13\%\\
      \hline
      Concession & 13\%\\
      \hline
      Asynchronous & 10\%\\
      \hline
      Level-of-detail & 9\%\\
      \hline
      Synchronous & 8\%\\
      \hline
      Instantiation & 4\%\\
      \hline
      Manner & 4\%\\
      \hline
      Progression & 3\%\\
      \hline
      Equivalence & 2\%\\
      \hline
      \textbf{total} & \textbf{86\%}\\
      \hline
\end{tabular}
\caption{ The most frequent Level-2 senses in our corpus}
 \end{center}
\end{table}

Obviously, there is approximately the same number of explicit and implicit relations in the corpus. This may indicate that explicit connectives and relations are more likely to present in Chinese spoken texts rather than Chinese written texts.\\

The figures shown by Table 4 illustrate the distributions of class level senses. We make a comparison for the class level senses between our corpus and the CUHK Discourse Treebank for Chinese (CUHK-DTBC). CUHK Discourse Treebank for Chinese is a corpus annotating news reports. Therefore, our comparison with it may shed light on the differences of discourse structures in different genres. According to the statistics of CUHK-DTBC for 400 documents and our corpus, while more than half of the senses is Expansion in CUHK-DTBC, it just represents 37.5\% in our corpus. In addition, it is highlighted that the ranks of the class level senses are the same in both corpora, although all of the other three senses in our corpus are more than those in CUHK-DTBC. \\

The most frequent second-level senses in our corpus can be seen from Table 5. We can find that 20\% of the senses is Cause (including Reason and Result), followed by Conjunction and Concession, each with 13\%. The top 10 most frequent senses take up 86\% of all senses annotated, which reveals that other senses also can validate their existence in our corpus. Therefore, these findings show that, compared with other corpora about Chinese shallow relations where the majority of the documents are news report, our corpus evidently show a more balanced and varied distribution from perspectives of both relations and senses, which in large measure proves the differences in discourse relations between Chinese written texts and Chinese spoken texts.
\\

\section{Conclusions and Future Work}

In this paper, we describe our scheme and process in annotating shallow discourse relations using PDTB-style. In view of the differences between English and Chinese, we made adaptations for the PDTB-3 scheme such as removing AltLexC and adding Progression into our sense hierarchy. To ensure the annotation quality, we formulated detailed annotation criteria and quality assurance strategies. After serious training, we annotated 3212 discourse relations, and we achieved a satisfactory consistency of labelling with a Kappa value of greater than 0.85 for most of the indicators. Finally, we display our annotation results in which the distribution of discourse relations and senses differ from that in other corpora which annotate news report or newspaper texts. Our corpus contains more Contingency, Temporal and Comparison relations instead of being governed by Expansion.\\

In future work, we are planning to 1) expand our corpus by annotating more TED talks or other spoken texts; 2) build a richer and diverse connective set and AltLex expressions set; 3) use the corpus in developing a shallow discourse parser for Chinese spoken discourses; 4) also explore automatic approaches for implicit discourse relations recognition. 

\section{Acknowledgement}
The present research was supported by the National Natural Science Foundation of China (Grant No. 61861130364) and the Royal Society (London) (NAF$\backslash$R1$\backslash$180122). We would like to thank the anonymous reviewers for their insightful comments.
\section{Bibliographical References}
\label{main:ref}

\bibliographystyle{lrec}
\bibliography{arxiv}

\end{CJK}
\end{document}